\documentclass{article} 
\usepackage{nips13submit_e,times}
\usepackage{fancyhdr} 
\usepackage{url}
\usepackage{graphicx}
\usepackage{epsfig}
\usepackage{amssymb}
\usepackage{epstopdf}
\usepackage{multirow}
\usepackage{booktabs}
\usepackage{threeparttable}
\usepackage{tabularx}
\usepackage[hidelinks]{hyperref}
\hypersetup{
    colorlinks,
    citecolor=green,
    filecolor=black,
    linkcolor=red,
    urlcolor=blue
}

\fancypagestyle{firststyle}
{
   
   \fancyhf{}
}
\title{Face Attribute Prediction Using Off-the-Shelf CNN Features}

\author{
Yang Zhong \qquad Josephine Sullivan \qquad Haibo Li \\
Computer Science and Communication\\
KTH Royal Institute of Technology\\
100 44 Stockholm, Sweden\\
\texttt{\{yzhong, sullivan, haiboli\}@kth.se} \\
}

\nipsfinalcopy 

\begin{document}

\maketitle

\begin{abstract}
Predicting attributes from face images in the wild is a challenging computer vision problem. 
To automatically describe face attributes from face containing images, traditionally one needs to cascade three technical blocks --- face localization, facial descriptor construction, and attribute classification --- in a pipeline. 
As a typical classification problem,  face attribute prediction has been addressed using deep learning. 
Current state-of-the-art performance was achieved by using two cascaded Convolutional Neural Networks (CNNs), which were specifically trained to learn face localization and attribute description. 
In this paper, we experiment with an alternative way of employing the power of deep representations from CNNs. 
Combining with conventional face localization techniques, we use off-the-shelf architectures trained for face recognition to build facial descriptors. 
Recognizing that the describable face attributes are diverse, our face descriptors are constructed from different levels of the CNNs for different attributes to best facilitate face attribute prediction. 
Experiments on two large datasets, LFWA and CelebA, show that our approach is entirely comparable to the state-of-the-art. 
Our findings not only demonstrate an efficient face attribute prediction approach, but also raise an important question: how to leverage the power of off-the-shelf CNN representations for novel tasks.  
\end{abstract}

\section{Introduction}

The recent success achieved by the Convolutional Neural Networks (CNNs) has vastly driven the advances in many aspects of computer vision, such as image classification and object detection, and pushed the boundaries of understanding image content through computer vision. 
In face recognition, we have witnessed great improvements brought by CNNs in solving the challenging large-scale face verification and recognition tasks \cite{Huang2007,2015arXiv151200596K}. 

Like recognizing identities, describing attributes from face images in the wild has been an active research topic for years.
Being able to automatically describe face attributes from face images in the wild is very challenging but can be very helpful.
For instance, one can not only build identifiers directly based on attributes \cite{kumar2011describable}, but also  efficiently construct highly flexible large-scale hierarchical datasets, which can further benefit image classification and attribute-to-image generation \cite{yanhd,yan2015attribute2image}.

\begin{figure}
\begin{center}
   \includegraphics[width=.7\linewidth]{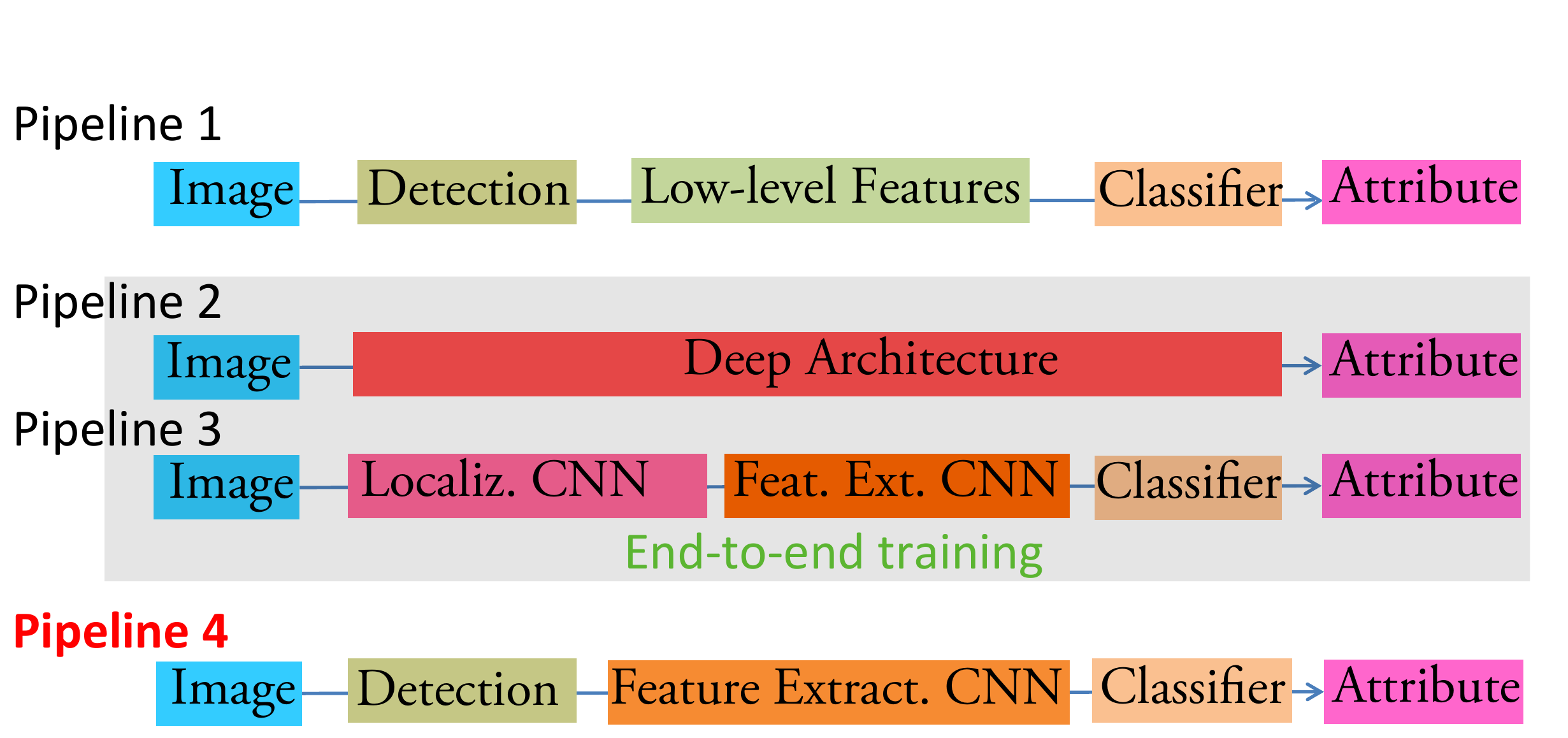}
\end{center}
\vspace{-.5em}
   \caption{Potential pipelines of automatic attribute estimation.}
\label{fig:intro}
\end{figure}

The general process of predicting face attributes is to construct face representations and train domain classifiers for prediction.  
As summarized in Figure \ref{fig:intro}, traditional approaches (Pipeline 1) construct low-level descriptors, such as SIFT \cite{lowe2004distinctive} and LBP \cite{ahonen2006face}, through landmark detection. 
These descriptors are then utilized for building attribute classifiers.
Similarly, by using CNNs one can also employ massive sentence and image training instances to construct end-to-end deep architectures (Pipeline 2) for learning semantic-visual correspondences as in \cite{karpathy2014deep,vinyals2014show}. 
However, such approaches are rather resource demanding. 
    
An intuitive alternative way (Pipeline 3) is to decompose the end-to-end network (by functionality) into a localization network, a feature construction network, and attribute classifiers, and build them individually as in \cite{liu2015faceattributes}. 
By cascading the trained components, such a pipeline can achieve state-of-the-art performance. 
However, the requirements of this approach on data and training efforts seem enormous.
In addition, it appears that to reach the best performance, high-level features must be used in the concatenated deep networks; 
fine-tuning on the pre-trained off-the-shelf high-level abstraction yields significantly better performance.

However, given that many face attributes are locally orientated and different layers of CNN features encode different levels of visual information, we believe face attributes would not be best represented by merely high-level features from deep neural networks. 
Thus, in this paper, we alternatively tackle the face attribute prediction problem using a pipeline composed of a conventional localization component, an off-the-shelf CNN, and attribute classifiers (Pipeline 4 in Figure \ref{fig:intro}).
Our focus is finding proper feature representations from pre-trained CNNs to boost attribute perdition.
We use off-the-shelf architectures and a publicly accessible model intended for face recognition to do feature construction, and investigate what types of feature representations from the network can efficiently improve face attribute prediction. 

Our investigations show that intermediate representations from pre-trained CNNs have distinct advantages over high-level features for the target face attribute prediction problem.  
By simply utilizing these features, we achieved very promising results on a par with the state-of-the-art, produced by the intensively trained two-stage CNN, on two recently released face attribute prediction datasets CelebA and LFWA \cite{liu2015faceattributes}.
Our findings also suggest that off-the-shelf intermediate CNN representations could be easily utilized when transferring from the source problem to novel detection and classification tasks.

\section{Related Work}
Traditionally, face descriptors were built from hand-crafted features. 
These features were constructed either from the whole face area, or extracted from detected local components and concatenated into a train of descriptors \cite{kumar2009attribute}.
Classifiers were trained based on these features to recognize the presence and quantitative degree of the domain attributes.
Recently, Liu.\ et\ al.\ \cite{liu2015faceattributes} proposed a cascaded learning framework to perform attribute prediction in the wild.
By pre-training and fine-tuning on large object dataset and face datasets, it efficiently localizes faces and produces semantic attributes for arbitrary face sizes without alignment.

As a strong feature learner, CNN has been successfully applied in face recognition, especially for solving the challenging face recognition in the wild problem \cite{Huang2007}. 
Besides the DeepID series approaches \cite{sun2013deep, sun2014deep, sun2015deepid3}, related efforts have also been made to pose correction \cite{taigman2013deepface},  architectures design \cite{szegedy2014going, schroff2015facenet} and data collection skills \cite{parkhi2015deep}.
With recently launched hardware platforms and the publicly accessible large-scale dataset \cite{yi2014learning}, developing deep learning based face recognition approaches becomes feasible with less resources.

\section{Attribute Prediction using CNNs Off-the-shelf}
\label{sec:exp}

\subsection{Overview}
To describe face appearances using CNN features, it is critical to first consider a proper face representation from the deep neural network.      
One natural way is to represent faces using the discriminatively learned features, from the high-level hidden layers, mostly used for representing identities in face recognition tasks, as in \cite{sun2013deep}. 
In this case, appearance attributes are embedded in the activation of neurons in the discriminative feature.

However, to describe the appearance using deep representations from CNNs, it is easy to expect that the selected representation should preserve the variability to describe the appearance variations regarding facial physical characteristics, such as ``big (eyes)'' and ``open (mouth)''.
While on the contrary, when attributes are identity correlated (e.g.\ gender and ethnicity), such representation should be robust with respect to non-identity related interference.
Thus, the representation that most suitable to describe a certain attribute highly depends on the property (e.g.\ if subject to identity) of the attribute itself. 
Given that a CNN enables its intermediate representations to maintain both discriminality and rich spatial information \cite{azizpour2014generic, razavian2014cnn}, it is therefore tenable to employ flexible selections of feature presentations for predicting face attributes. 

\subsection{Experiments}
\subsubsection{Procedures}
To identify the most effective deep representations, our method explores the attribute prediction power of intermediate representations versus the final representation \footnote{The high level abstraction used for representing identity, which is often extracted from the last FC layer.} from CNNs trained for face recognition.
Therefore, we first trained a face classification CNN (or use a publicly available model), 
then we evaluated the prediction performance of the representations extracted from different levels of the CNN.  
The training of CNNs and the evaluation of prediction power were conducted separately on two independent datasets: the WebFace \cite{yi2014learning} was used for CNN training, and the CelebA and LFWA for evaluation.
We used two well known off-the-shelf architectures (\emph{configurations of filter stacks}) in our experiments to  benefit from the latest development in CNN architecture design.

\textbf{Network architecture:} 
The networks used in our experiments shared the same format: they were composed of off-the-shelf filter stacks followed by two Fully Connected (denoted by $FC1$ and $FC2$) layers.
Considering the ease of training and efficient inference during test phase, we selected Google's \textit{FaceNet NN.1} \cite{schroff2015facenet} (shortened as ``FaceNet'' in the following) and VGG's \textit{``very deep'' model} \cite{simonyan2014very} as the structure of convolutional (conv.) layers.
The CNNs were trained in the most fundamental flat classification manner with a Softmax layer attached to the last FC layer during training.
We used dropout regularization between FCs to prevent overfitting and the dropout rate was set to 0.5 for all FC layers in our networks. 
PReLU\cite{he2015delving} rectification was attached to each convolution and FC layer.

\textbf{Training:} 
Around $10,000$ identities with $350,000$ image instances of the WebFace dataset were used. 
Random mirroring, slight rotation and jittering were utilized as data augmentation. 
The learning rate was initially set to $0.015$, and then decreased by a factor of 10 when the validation set accuracy stopped increasing. 
The networks were trained by 3 decreasing learning rates.
Faces were segmented and normalized to a size of $120\times120$ and randomly cropped patches of $112\times112$ were fed into the network.

\begin{figure}
\begin{center}
   \includegraphics[width=0.6\linewidth]{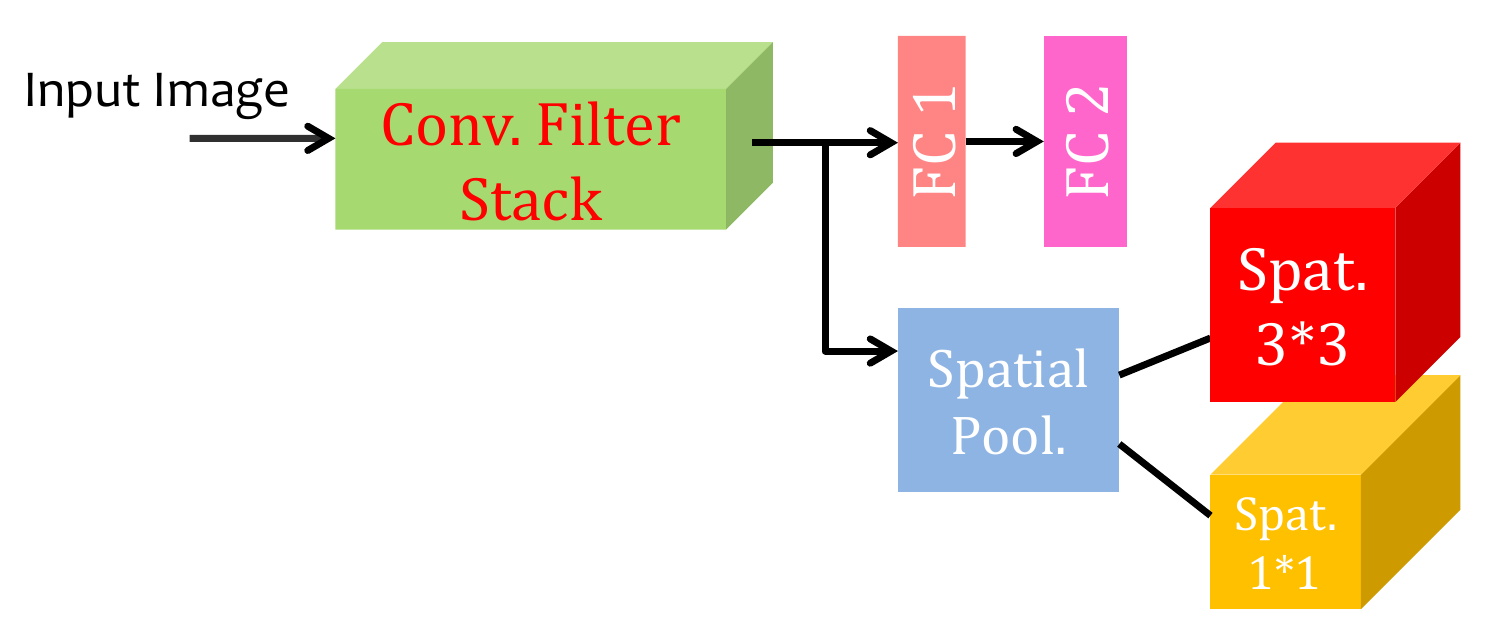}
\end{center}
   \caption{Pipeline of extracting deep representations from trained CNN. Intermediate features ($Spat. N \times N$, FC1) and final representation FC2 are extracted from the trained network. $N$ for the side of deep feature map after extra pooling step. In total, 4 types of representations will be studied for face attribute prediction.}
\label{fig:repExtract}
\end{figure}

\textbf{Feature Extraction:} 
To extract face descriptors from CNNs, only the center patch ($112\times112$) and its mirrored version of aligned face images were fed into the CNNs unless otherwise stated. 
We aligned faces using feature points detected by random forests \cite{kazemi2014one}.
We took the averaged representations of the two fed-in patches at different levels of the network, i.e.\ ``$Spat.1 \times 1$'', ``$Spat.3\times 3$'', ``$FC 1$'', and ``$FC 2$'' , as shown in Figure \ref{fig:repExtract}, and evaluated their attribute estimation performance to identify the most effective representation corresponding to each attribute. 

The output of the last conv. filter stack was selected as the representative of the intermediate representations since it was shown to have the most discriminality and spatial information for recognition and image retrieval \cite{azizpour2014generic}.
Extra max. pooling steps were applied to reduce the dimension of intermediate spatial representations.
Then $Spat.3\times 3$ and $Spat.1\times 1$ were of $3 \times 3 \times K $ and $1 \times 1 \times K $ in our experiments regardless of the network, where $K$ represents the channel depth of the employed network.

\textbf{Attribute prediction: }
The face attribute prediction performance was evaluated on the released version of CelebA and LFWA datasets\footnote{\url{http://mmlab.ie.cuhk.edu.hk/projects/CelebA.html}. Released in Oct. 2015.}.
The CelebA contains approximately $200,000$ images of $10,000$ identities and LFWA has $13,233$ images of $5,749$ identities. 
Each image in CelebA and LFWA is annotated with $40$ binary attribute tags.  
We used the same procedure to build our attribute classifier as in \cite{liu2015faceattributes}:
binary linear SVM \cite{fan2008liblinear} classifiers were trained directly for all levels of representations (i.e.\ $FCs$ and $Spat.'s$) to classify face attributes.
On the CelebA, the training set for each attribute classifier had $20,000$ image instances (where available). 
Since this dataset and the data for training our CNN are independent (the learning targets are also different), we tested the attribute prediction accuracy of our classifiers across the whole dataset through random selection of training and testing face instances.
On the LFWA, we took the training instances defined by the dataset.  
We report the prediction accuracy as the mean of True Acceptance Rate and True Rejection Rate for each attribute on both CelebA and LFWA datasets.

\textbf{Evaluations and Comparison: }
The same evaluation protocol as in \cite{liu2015faceattributes} was used in our experiments.
Since the features in our experiment was extracted from aligned face images and the alignment process was independent of the network, we selected the corresponding approach (``[17]+ ANet'' in \cite{liu2015faceattributes}) as the \textbf{baseline method}. 
The current state of the art in \cite{liu2015faceattributes} is denoted by ``Two-stage CNN'' and ``LNet+ANet'' in this paper. 

The above mentioned procedures were used in the following experiments.
We first employed our FaceNet to thoroughly study the discrepancy between different representation types for face attribute prediction. 
The identified best performing off-the-shelf features were utilized to challenge on the CelebA and the LFWA to compare with \cite{liu2015faceattributes}.
We then extended our experiments by further investigating different configurations of the FaceNet, the VGG's ``very deep'' architecture, and the publicly available VGG-Face model\footnote{\url{http://www.robots.ox.ac.uk/~vgg/software/vgg_face/}, accessed in Nov. 2015.} to ensure the discrepancy in attribute prediction power among the deep representations. 

\subsubsection{Performance Discrepancy between Deep Representations}
\begin{figure*}[]
\begin{center}
   \includegraphics[width=1\linewidth]{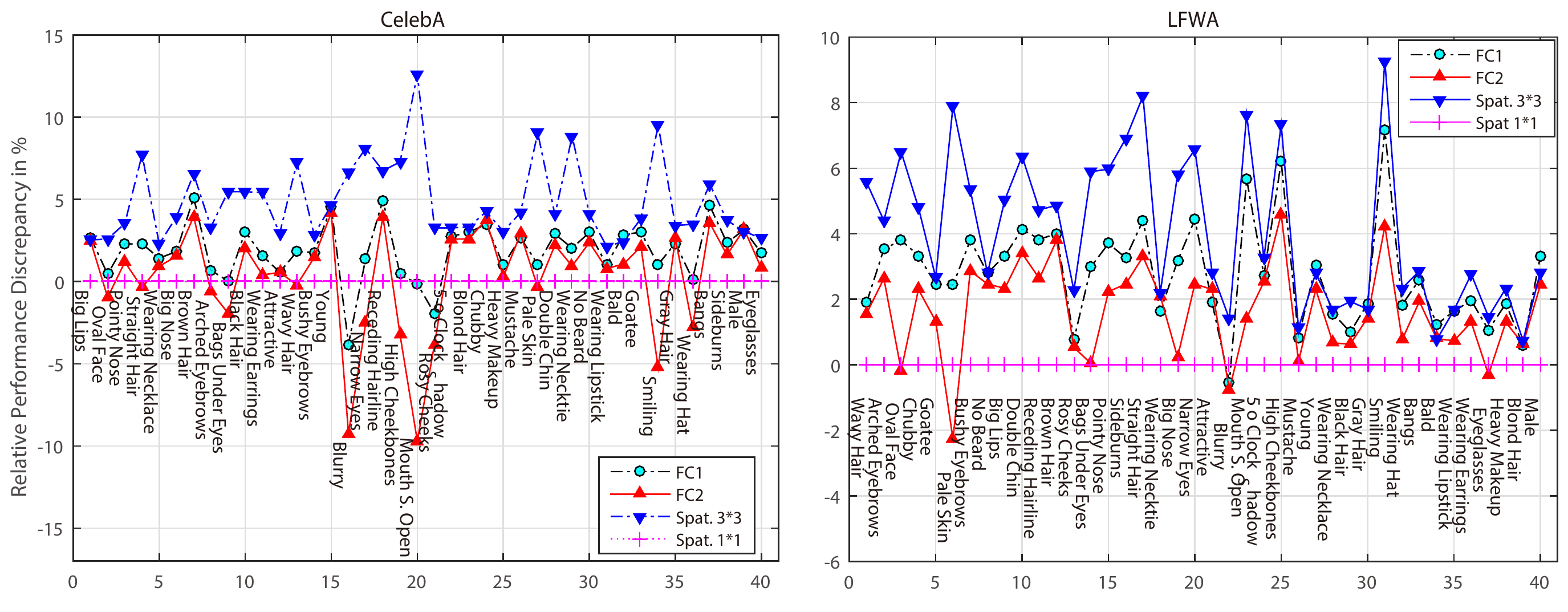}
\end{center}
\vspace{-1em}
   \caption{Comparing prediction performance of deep representations on CelebA and LFWA. $Spat. 1 \times 1$ was set as the reference for each attribute and the relative prediction power of each representation type is plotted based on its difference to the reference. The attributes are sorted based on the absolute prediction accuracy of the baseline method on each dataset. On CelebA, the \textbf{absolute mean prediction accuracy} of $FC1=83\%$ , $FC2=82\%$ ,  $Spat.3 \times 3=86\%$ , $Spat.1 \times 1=82\%$, and on LFWA, $FC1=83\%$ , $FC2=82\%$ ,  $Spat.3 \times 3=85\%$ , $Spat.1 \times 1=81\%$. }
\label{fig:perfDec}
\end{figure*}

Our intuition as stated above was that the intermediate face representations would be more suitable for describing diverse types of attributes regarding their physical characteristics and image conditions.
To validate this, we trained a face recognition CNN with a structure of FaceNet.
The length of both FC layers was set to $512$ to reduce the risk of overfitting. 
The recognition rate of the trained FaceNet on the validation set was less than 98\% and the face verification performance on the LFW \cite{Huang2007} was 97.5\%.
We then extracted the four types of face representations, $Spat.1b1$, $Spat.3b3$, $FC1$, and $FC2$, from our trained model and linear classifiers were constructed and evaluated respectively on the training set.
The prediction performance for each representation type on all attributes is shown in Figure \ref{fig:perfDec}.

While it is intuitive that $FC2$, the identity discriminative feature, is unlikely to be the best choice for describing facial attributes all the time, it is still astonishing that $FC2$ was significantly outperformed ($\geqslant5\%$ in prediction accuracy) by others on $13$ attributes. 
Similar disadvantages can also be observed on the LFWA dataset.
It is easy to find that: 
\begin{enumerate}  
\item Representations at different levels of the network feature quite diverse performance in attribute description.  
\item Intermediate representations, especially $Spat.3 \times 3$, are likely more effective in telling the weak identity-related attributes describing expressions and image conditions, which counts more on spatial information. 
\end{enumerate}
For instance, for attributes related to mouth and eyes which can produce dynamic facial expressions, such performance gaps are significant.
For ``$Spat.3 \times 3$'' which better preserved spatial information, it is more effective in specifying shape and motion of facial components (e.g.\ Attribute 20 and 34 on CelebA in Figure \ref{fig:perfDec}).
This is natural since intermediate representations contain mid-level features composed by low-level ones, thus they are more suitable to describe local facial attributes.

\begin{figure*}
\begin{center}
   \includegraphics[width=1\linewidth]{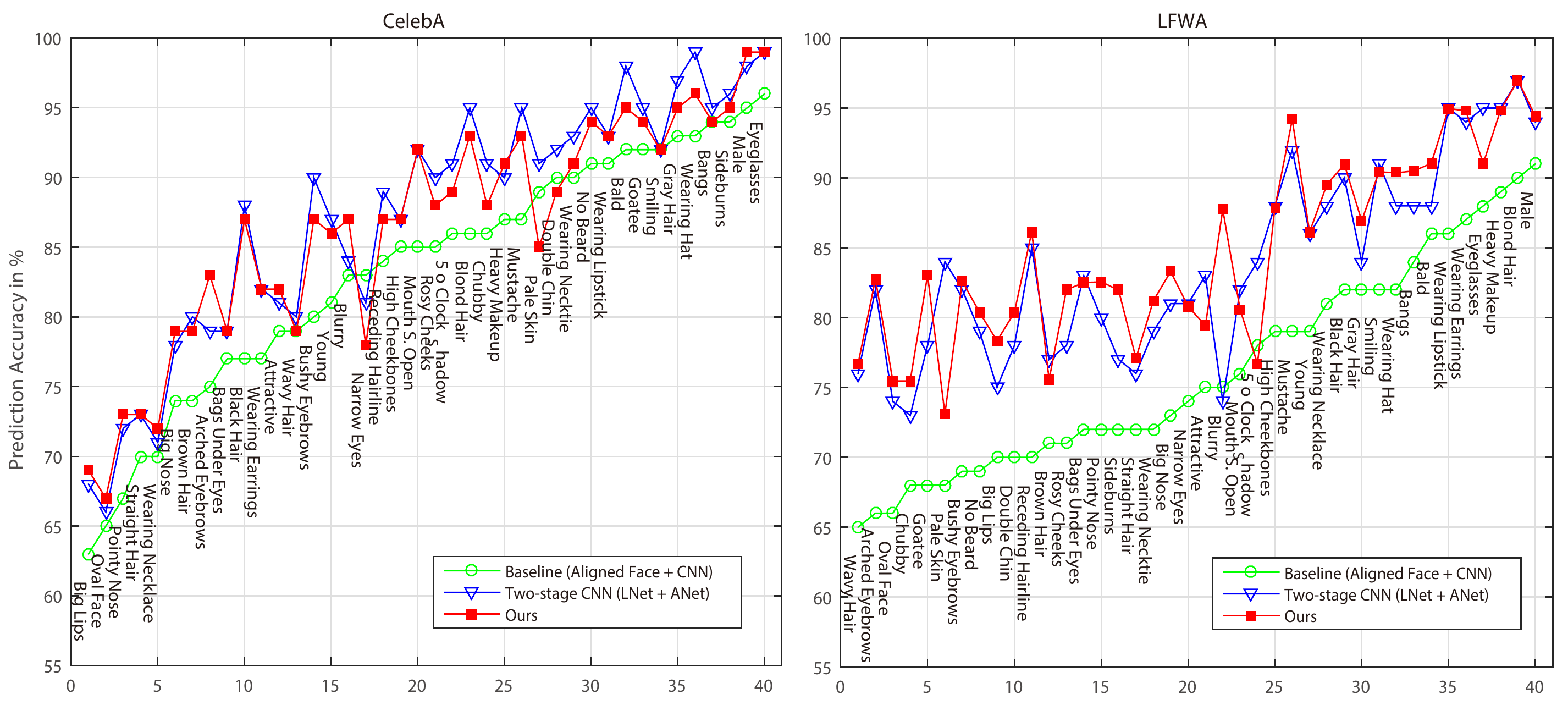}
\end{center}
   \caption{Comparing attribute prediction results on CelebA and LFWA.}
\label{fig:resCombine}
\end{figure*}

Our investigations show that the best performing representations achieved attribute prediction accuracy of 86.6\% on CelebA and 84.7\% on LFWA, which is on a par with state-of-the-art ``Two-stage CNN'' approach which was trained with massive image classification and face data. 
The comparative results are listed in Table \ref{tabFinal} and shown in Figure \ref{fig:resCombine}. 
One can see that:
\begin{enumerate}
\item By leveraging the intermediate deep representations from various levels of CNNs, the equivalent baseline approach is outperformed with a big margin. 
\item Even without fine-tuning the pre-trained CNN, our average prediction performance is still comparable to the state-of-the-art on both datasets.
\end{enumerate}

Here we noticed that the intermediate representations dominated the best representations of the attributes.
This indicates that spatial information, i.e.\ location and magnitude of activation in conv.\ filter responses, is significant for describing attributes;
if one wants to utilize high-level features, which implicitly embeds spatial information, fine-tuning must be conducted on the high-level abstractions to enhance such useful spatial information.

\begin{table*}[]
\centering
\tiny
\caption{Comparing prediction accuracy (in \%) on CelebA and LFWA: corresponding average values of our approaches are \textbf{86.6\%} and \textbf{84.7\%}; for the baseline method: 83\%, 76\%; for the current best LNet+ANet approach: \textit{87\%} and \textit{84\%}.}
\label{tabFinal}
\begin{tabular}{|l|c|c|c|c|c|c|c|c|c|c|c|c|c|c|c|c|c|c|c|c|c|}
 \toprule
 \hline
\multicolumn{2}{|c|}{}                                   & \rotatebox[origin=c]{90}{5 o Clock Shadow}       & \rotatebox[origin=c]{90}{Arched Eyebrows}         & \rotatebox[origin=c]{90}{Attractive}              & \rotatebox[origin=c]{90}{Bags Under Eyes}         & \rotatebox[origin=c]{90}{Bald}                    & \rotatebox[origin=c]{90}{Bangs}                   & \rotatebox[origin=c]{90}{Big Lips}               & \rotatebox[origin=c]{90}{Big Nose}                & \rotatebox[origin=c]{90}{Black Hair}              & \rotatebox[origin=c]{90}{Blond Hair}             & \rotatebox[origin=c]{90}{Blurry}                  & \rotatebox[origin=c]{90}{Brown Hair}              & \rotatebox[origin=c]{90}{Bushy Eyebrows}          & \rotatebox[origin=c]{90}{Chubby}                  & \rotatebox[origin=c]{90}{Double Chin}             & \rotatebox[origin=c]{90}{ Eyeglasses }             & \rotatebox[origin=c]{90}{Goatee }                 & \rotatebox[origin=c]{90}{Gray Hair}               & \rotatebox[origin=c]{90}{Heavy Makeup}            & \rotatebox[origin=c]{90}{High Cheekbones}         \\ \hline
\multirow{3}{*}{\rotatebox[origin=c]{90}{CelebA}} & Baseline                       & 86                      & 75                      & 79                      & 77                      & 92                      & 94                      & 63                      & 74                      & 77                      & 86                      & 83                      & 74                      & 80                      & 86                      & 90                      & 96                      & 92                      & 93                      & 87                      & 85                      \\ \cline{2-22} 
                        & LNet+ANet                      & 91                      & 79                      & 81                      & 79                      & 98                      & 95                      & 68                      & 78                      & 88                      & 95                      & 84                      & 80                      & 90                      & 91                      & 92                      & 99                      & 95                      & 97                      & 90                      & 87                      \\ \cline{2-22} 
                        & Ours                           & 89                      & 83                      & 82                      & 79                      & 96                      & 94                      & 70                      & 79                      & 87                      & 93                      & 87                      & 79                      & 87                      & 88                      & 89                      & 99                      & 94                      & 95                      & 91                      & 87                      \\ \hline
\multirow{3}{*}{\rotatebox[origin=c]{90}{LFWA}}   & \multicolumn{1}{l|}{Baseline}  & \multicolumn{1}{l|}{78} & \multicolumn{1}{l|}{66} & \multicolumn{1}{l|}{75} & \multicolumn{1}{l|}{72} & \multicolumn{1}{l|}{86} & \multicolumn{1}{l|}{84} & \multicolumn{1}{l|}{70} & \multicolumn{1}{l|}{73} & \multicolumn{1}{l|}{82} & \multicolumn{1}{l|}{90} & \multicolumn{1}{l|}{75} & \multicolumn{1}{l|}{71} & \multicolumn{1}{l|}{69} & \multicolumn{1}{l|}{68} & \multicolumn{1}{l|}{70} & \multicolumn{1}{l|}{88} & \multicolumn{1}{l|}{68} & \multicolumn{1}{l|}{82} & \multicolumn{1}{l|}{89} & \multicolumn{1}{l|}{79} \\ \cline{2-22} 
                        & \multicolumn{1}{l|}{LNet+ANet} & \multicolumn{1}{l|}{84} & \multicolumn{1}{l|}{82} & \multicolumn{1}{l|}{83} & \multicolumn{1}{l|}{83} & \multicolumn{1}{l|}{88} & \multicolumn{1}{l|}{88} & \multicolumn{1}{l|}{75} & \multicolumn{1}{l|}{81} & \multicolumn{1}{l|}{90} & \multicolumn{1}{l|}{97} & \multicolumn{1}{l|}{74} & \multicolumn{1}{l|}{77} & \multicolumn{1}{l|}{82} & \multicolumn{1}{l|}{73} & \multicolumn{1}{l|}{78} & \multicolumn{1}{l|}{95} & \multicolumn{1}{l|}{78} & \multicolumn{1}{l|}{84} & \multicolumn{1}{l|}{95} & \multicolumn{1}{l|}{88} \\ \cline{2-22} 
                        & Ours                           & \multicolumn{1}{l|}{77} & \multicolumn{1}{l|}{83} & \multicolumn{1}{l|}{79} & \multicolumn{1}{l|}{83} & \multicolumn{1}{l|}{91} & \multicolumn{1}{l|}{91} & \multicolumn{1}{l|}{78} & \multicolumn{1}{l|}{83} & \multicolumn{1}{l|}{90} & \multicolumn{1}{l|}{97} & \multicolumn{1}{l|}{88} & \multicolumn{1}{l|}{76} & \multicolumn{1}{l|}{83} & \multicolumn{1}{l|}{75} & \multicolumn{1}{l|}{80} & \multicolumn{1}{l|}{91} & \multicolumn{1}{l|}{83} & \multicolumn{1}{l|}{87} & \multicolumn{1}{l|}{95} & \multicolumn{1}{l|}{88} \\ \hline
\multicolumn{22}{|l|}{}                                                                                                                                                                                                                                                                                                                                                                                                                                                                                                                                                                          \\ \hline
\multicolumn{2}{|l|}{}                                   & \rotatebox[origin=c]{90}{Male}                    & \rotatebox[origin=c]{90}{Mouth S. Open}           & \rotatebox[origin=c]{90}{Mustache}                & \rotatebox[origin=c]{90}{Narrow Eyes}             & \rotatebox[origin=c]{90}{No Beard}                & \rotatebox[origin=c]{90}{Oval Face}               & \rotatebox[origin=c]{90}{Pale Skin}               & \rotatebox[origin=c]{90}{Pointy Nose}             & \rotatebox[origin=c]{90}{Receding Hairline}       & \rotatebox[origin=c]{90}{Rosy Cheeks}             & \rotatebox[origin=c]{90}{Sideburns}               & \rotatebox[origin=c]{90}{Smiling}                 & \rotatebox[origin=c]{90}{Straight Hair}           & \rotatebox[origin=c]{90}{Wavy Hair}               &\rotatebox[origin=c]{90}{ Wearing Earrings }       & \rotatebox[origin=c]{90}{Wearing Hat}             & \rotatebox[origin=c]{90}{Wearing Lipstick}       & \rotatebox[origin=c]{90}{Wearing Necklace}        & \rotatebox[origin=c]{90}{Wearing Necktie}         & \rotatebox[origin=c]{90}{Young}                   \\ \hline
\multirow{3}{*}{\rotatebox[origin=c]{90}{CelebA}} & Baseline                       & 95                      & 85                      & 87                      & 83                      & 91                      & 65                      & 89                      & 67                      & 84                      & 85                      & 94                      & 92                      & 70                      & 79                      & 77                      & 93                      & 91                      & 70                      & 90                      & 81                      \\ \cline{2-22} 
                        & LNet+ANet                      & 98                      & 92                      & 95                      & 81                      & 95                      & 66                      & 91                      & 72                      & 89                      & 90                      & 96                      & 92                      & 73                      & 80                      & 82                      & 99                      & 93                      & 71                      & 93                      & 87                      \\ \cline{2-22} 
                        & Ours                           & 99                      & 92                      & 93                      & 78                      & 94                      & 67                      & 85                      & 73                      & 87                      & 88                      & 95                      & 92                      & 73                      & 79                      & 82                      & 96                      & 93                      & 73                      & 91                      & 86                      \\ \hline
\multirow{3}{*}{\rotatebox[origin=c]{90}{LFWA}}   & \multicolumn{1}{l|}{Baseline}  & \multicolumn{1}{l|}{91} & \multicolumn{1}{l|}{76} & \multicolumn{1}{l|}{79} & \multicolumn{1}{l|}{74} & \multicolumn{1}{l|}{69} & \multicolumn{1}{l|}{66} & \multicolumn{1}{l|}{68} & \multicolumn{1}{l|}{72} & \multicolumn{1}{l|}{70} & \multicolumn{1}{l|}{71} & \multicolumn{1}{l|}{72} & \multicolumn{1}{l|}{82} & \multicolumn{1}{l|}{72} & \multicolumn{1}{l|}{65} & \multicolumn{1}{l|}{87} & \multicolumn{1}{l|}{82} & \multicolumn{1}{l|}{86} & \multicolumn{1}{l|}{81} & \multicolumn{1}{l|}{72} & \multicolumn{1}{l|}{79} \\ \cline{2-22} 
                        & \multicolumn{1}{l|}{LNet+ANet} & \multicolumn{1}{l|}{94} & \multicolumn{1}{l|}{82} & \multicolumn{1}{l|}{92} & \multicolumn{1}{l|}{81} & \multicolumn{1}{l|}{79} & \multicolumn{1}{l|}{74} & \multicolumn{1}{l|}{84} & \multicolumn{1}{l|}{80} & \multicolumn{1}{l|}{85} & \multicolumn{1}{l|}{78} & \multicolumn{1}{l|}{77} & \multicolumn{1}{l|}{91} & \multicolumn{1}{l|}{76} & \multicolumn{1}{l|}{76} & \multicolumn{1}{l|}{94} & \multicolumn{1}{l|}{88} & \multicolumn{1}{l|}{95} & \multicolumn{1}{l|}{88} & \multicolumn{1}{l|}{79} & \multicolumn{1}{l|}{86} \\ \cline{2-22} 
                        & \multicolumn{1}{l|}{Ours}      & \multicolumn{1}{l|}{94} & \multicolumn{1}{l|}{81} & \multicolumn{1}{l|}{94} & \multicolumn{1}{l|}{81} & \multicolumn{1}{l|}{80} & \multicolumn{1}{l|}{75} & \multicolumn{1}{l|}{73} & \multicolumn{1}{l|}{83} & \multicolumn{1}{l|}{86} & \multicolumn{1}{l|}{82} & \multicolumn{1}{l|}{82} & \multicolumn{1}{l|}{90} & \multicolumn{1}{l|}{77} & \multicolumn{1}{l|}{77} & \multicolumn{1}{l|}{94} & \multicolumn{1}{l|}{90} & \multicolumn{1}{l|}{95} & \multicolumn{1}{l|}{90} & \multicolumn{1}{l|}{81} & \multicolumn{1}{l|}{86} \\ \hline
\end{tabular}
\end{table*}


\subsubsection{Further Validations} 
To further verify the potential utility of intermediate spatial representations for face attribute prediction, we also evaluated various network architectures trained by different configurations, 
which for each model are listed in Table \ref{tab:netconfig}. 
CelebA was selected as the evaluation dataset due to its larger scale.

Specifically, we first evaluated two different networks of the FaceNet architecture.
Model 1 was trained by the first $8,000$ identities that has the most images on the WebFace dataset (i.e.\ taking away the long-tail data). 
Since the length of the representing features plays a vital role in face representation \cite{parkhi2015deep}, we then evaluated the influences of varying FC layer lengths in face attribute prediction with Model 2 by increasing the length of FC layers to $1024$. 
The receptive field was kept the same ($112 \times 112$).

We also cross-validated the utility of the deep representations with VGG type architectures in Model 3 and 4.
Model 3 had filter stack as VGG fitler-Config.C \cite{simonyan2014very}, but with a duplicated conv.\ and pooling section appended to the fifth pooling so that it was even deeper.
(Thus, for this configuration, the filter stack directly gave output with size of $3 \times 3$. 
It was then max. pooled to get $1 \times 1$ output.)
To bring more divergence, we decreased the input size to $96 \times 96$ (still cropped from $120 \times 120$) and set FCs to $4096$. 
Model 4 was the off-the-shelf \textit{VGG-Face} network. 
The receptive area for Model 4 was $224 \times 224$.
The corresponding results in terms of the averaged prediction accuracy are provided in Table \ref{tab:CrosVali}.

\begin{table}[]
\centering
\footnotesize
\caption{CNNs and training data used for further validations.}
\label{tab:netconfig}
\begin{tabular}{ccccc}
\hline \toprule
Model & \begin{tabular}[c]{@{}c@{}}Conv. Filter Stack\\ Architecture\end{tabular} & \begin{tabular}[c]{@{}c@{}}FC 1\\ Dim.\end{tabular} & \begin{tabular}[c]{@{}c@{}}FC 2\\ Dim.\end{tabular} & \begin{tabular}[c]{@{}c@{}}Training Dataset / \\ Identities\end{tabular} \\ \hline
1     & FaceNet                                                        & 512                                                 & 512                                                & WebFace, ~8k \\ \hline
2     & FaceNet                                                        & 1024                                                & 1024                                               & WebFace, ~10k  \\ \hline
3     & VGG, Config. C                                                        & 4096                                                & 4096                                               & WebFace, ~10k                                                \\ \hline
4     & VGG-Face                                                       & 4096                                                & 4096                                               &  private, \textgreater2.6k  \\  \bottomrule
\end{tabular}
\end{table}
\begin{table}[]
\footnotesize
\centering
\caption{ Decomposition of the best representations of the architectures in Table \ref{tab:netconfig}. This table gives the number of each representation type that formed the best representation for each model and provides the average prediction accuracy achieved by the best representations and the $FC2$ representation.  (``S.'' for ``Spat.'')}
\label{tab:CrosVali}
\begin{tabular}{ccccccc}
\hline \toprule
\multirow{2}{*}{Model} & \multicolumn{4}{c}{\# of Best Rep. from} & \multicolumn{2}{c}{ave. accuracy} \\ \cline{2-7} 
                  & S.3*3          & S.1*1          & FC1          & FC2         & Best Rep.      & FC2          \\ \hline
1			      & 33              & 0               & 6            & 1           & 86\%        & 84\%      \\ \hline
2			      & 31              & 0               & 6            & 3           & 86\%        & 84\%      \\ \hline
3      			  & 28              & 11              & 1            & 0           & 85\%        & 83\%      \\ \hline
4                 & 37              & 1               & 0            & 2           & 86\%        & 85\%      \\  \bottomrule
\end{tabular}
\end{table}

We observed that on average the spatial representations excelled on more than $75\%$ of the $40$ attributes.
The spatial representation from the off-the-shelf VGG-Face model even dominated the best representations. 
We attribute it to the dramatic increase of the receptive area. 
The intermediate representations embedded more detailed spatial information also further boosted the performance of $FC2$, which was as effective as the $Spat. 3\times3$.
The slightly worse performance of the features from Model 3 can be attributed to the lower receptive field and the 6th extra pooling, which caused transfer of prediction power from $Spat. 3 \times 3$ to $Spat. 1\times1$. 

Through further analysis of the results, we found that the intermediate spatial representations predicted 5 attributes (``Bags Under Eyes'' , ``Blurry'', ``Mouth S. Open'', ``Pale Skin'' and ``Narrow Eyes'') much better than the last FC representations. 

We believe the reason intermediate spatial representations outperformed on so many attributes is that these \textbf{human describable} attributes are more likely to be identified from the spatial information captured by human brains.
Considering these attributes are semantic concepts relating to specific domains and these domains by themselves alone can hardly be used to pin-point a specific identity from a crowd of people, the utility of the intermediate features, which are less discriminating than the high level abstraction, from CNNs makes sense.

\section{Conclusions}
\label{conclusions}





In this paper, we address the problem of predicting face attributes using CNNs trained for face recognition.
We employ CNNs with off-the-shelf architectures and publicly available models --- Google's FaceNet and VGG's ``very deep'' model --- with the conventional pipeline to study the prediction power of different representations from the trained CNNs.
Our investigations present the correspondence diversity between the best performing representations and the human describable attributes.
They also reveal that the intermediate representations from CNNs are very effective in predicting facial attributes in general. 
Although previous works have shown that fine-tuning the pre-trained networks brought significant improvements when transferring to novel tasks, we empirically demonstrate that intermediate deep features from pre-trained networks can also form a promising alternative. 
By properly leveraging these off-the-shelf CNN representations, we achieved accurate attribute prediction on a par with current state-of-the-art performance.

\section*{Acknowledgments}
We gratefully acknowledge the support from NVIDIA Corporation for GPU donations. 
We have enjoyed discussions with Ali Sharif Razavian and Atsuto Maki.


\bibliographystyle{plain}
\bibliography{ref.bib}

\end{document}